\documentclass[10pt]{article}
\usepackage{amsmath,amsthm,amsfonts,amssymb,latexsym,graphicx}
\usepackage[noadjust]{cite}
\usepackage{algorithm}
\usepackage[noend]{algpseudocode}
\algrenewcommand\algorithmicthen{\relax}
\algrenewcommand\algorithmicdo{\relax}
\usepackage[pdfpagemode=UseNone,pdfstartview=FitH]{hyperref}

\newcommand{\OCMIV}{Fedorova/etal:2012ICML}
\newcommand{\OCMXXIV}{Vovk:2021}
\newcommand{\OCMXXIX}{Vovk:arXiv2006}
\newcommand{\OCMXXXII}{Vovk/etal:arXiv2102}

\allowdisplaybreaks[4]

\theoremstyle{definition}

\newtheorem{remark}{Remark}

\newcommand*{\ddd}{\mathrm{d}}

\title{Conformal testing in a binary model situation}
\author{Vladimir Vovk}

\begin{document}
\maketitle
\begin{abstract}
  Conformal testing is a way of testing the IID assumption based on conformal prediction.
  The topic of this note is computational evaluation of the performance of conformal testing
  in a model situation in which IID binary observations generated from a Bernoulli distribution
  are followed by IID binary observations generated from another Bernoulli distribution,
  with the parameters of the distributions and changepoint unknown.
  Existing conformal test martingales can be used for this task and work well in simple cases,
  but their efficiency can be improved greatly.

  The version of this note at \url{http://alrw.net} (Working Paper 33)
  is updated most often.
\end{abstract}

\section{Introduction}

The method of conformal prediction \cite[Chapter 2]{Vovk/etal:2005book}
can be adapted to testing the IID model \cite[Section 7.1]{Vovk/etal:2005book}.
For a long time it had remained unclear how efficient conformal testing is,
but \cite[Section 6]{\OCMXXIV} argued that in the binary case conformal testing
is efficient at least in a crude sense.
This note confirms that claim using simulation studies in a simple model situation.

The usual testing procedures in mathematical statistics \cite{Lehmann/Romano:2005}
are performed in the batch mode:
we are looking for evidence against the null hypothesis when given a batch of data
(a dataset of observations).
Conformal testing processes the observations sequentially (online),
and the amount of evidence found against the null hypothesis is updated when new observations arrive.
Valid testing procedures are equated with \emph{test martingales},
i.e., nonnegative processes with initial value 1 that are martingales under the null hypothesis.
Online hypothesis testing has been promoted in, e.g.,
\cite{Shafer/Vovk:2019,Shafer:arXiv1903,Grunwald/etal:arXiv1906,Ramdas/etal:arXiv2102}.

Our simulation studies will explore the performance of various test martingales,
including conformal test martingales (as defined in, e.g., \cite{\OCMXXXII}).
Conformal prediction uses randomization for tie-breaking,
and this feature is inherited by conformal testing.
In particular, conformal test martingales are randomized.
All plots in this note have been produced using the seed 0 for the NumPy random number generator,
and the dependence on the seed does not change any of our conclusions.

\begin{remark}
  In this note I will avoid the expression ``conformal martingale'',
  as used in \cite{\OCMXXIV},
  to avoid terminology clash with the notion of conformal martingale
  introduced in \cite{Getoor/Sharpe:1972} and discussed in \cite{Walsh:1977}.
  (Even though this would not have led to any confusion;
  the two notion are unlikely to be used in the same context.)
\end{remark}

\section{Model situation}

\begin{figure}
  \begin{center}
    \includegraphics[width=0.48\textwidth]{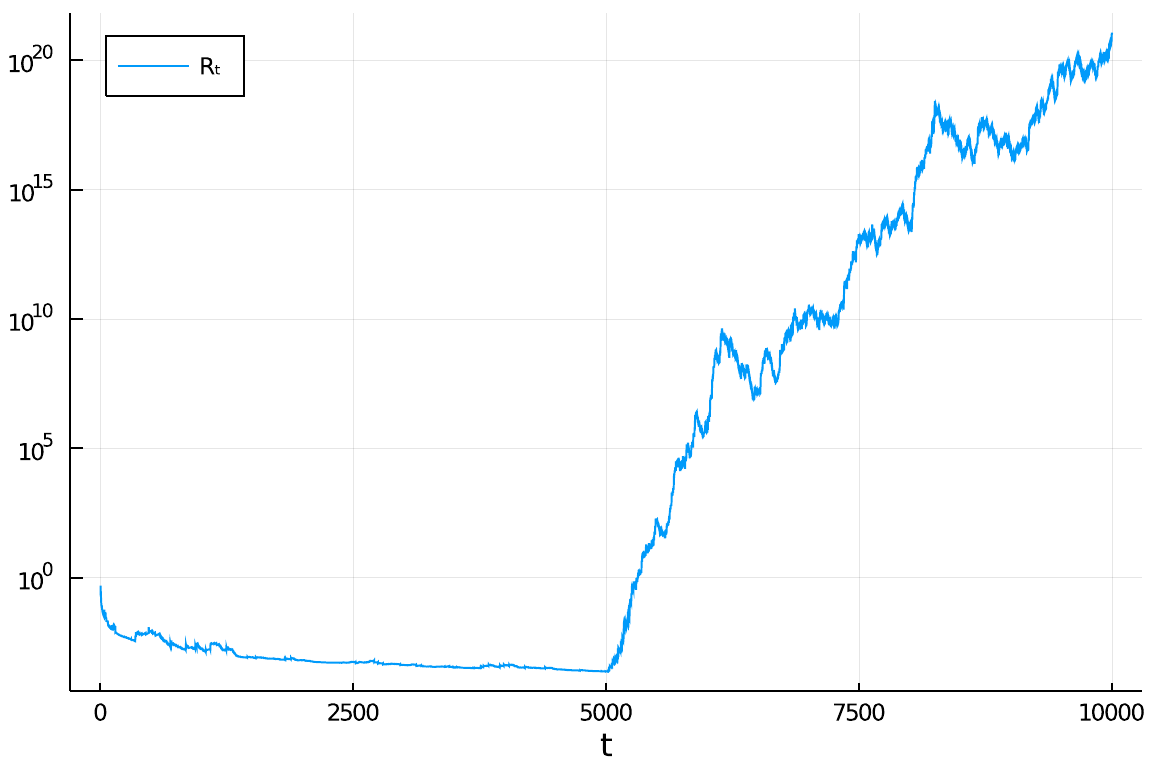}
    \includegraphics[width=0.48\textwidth]{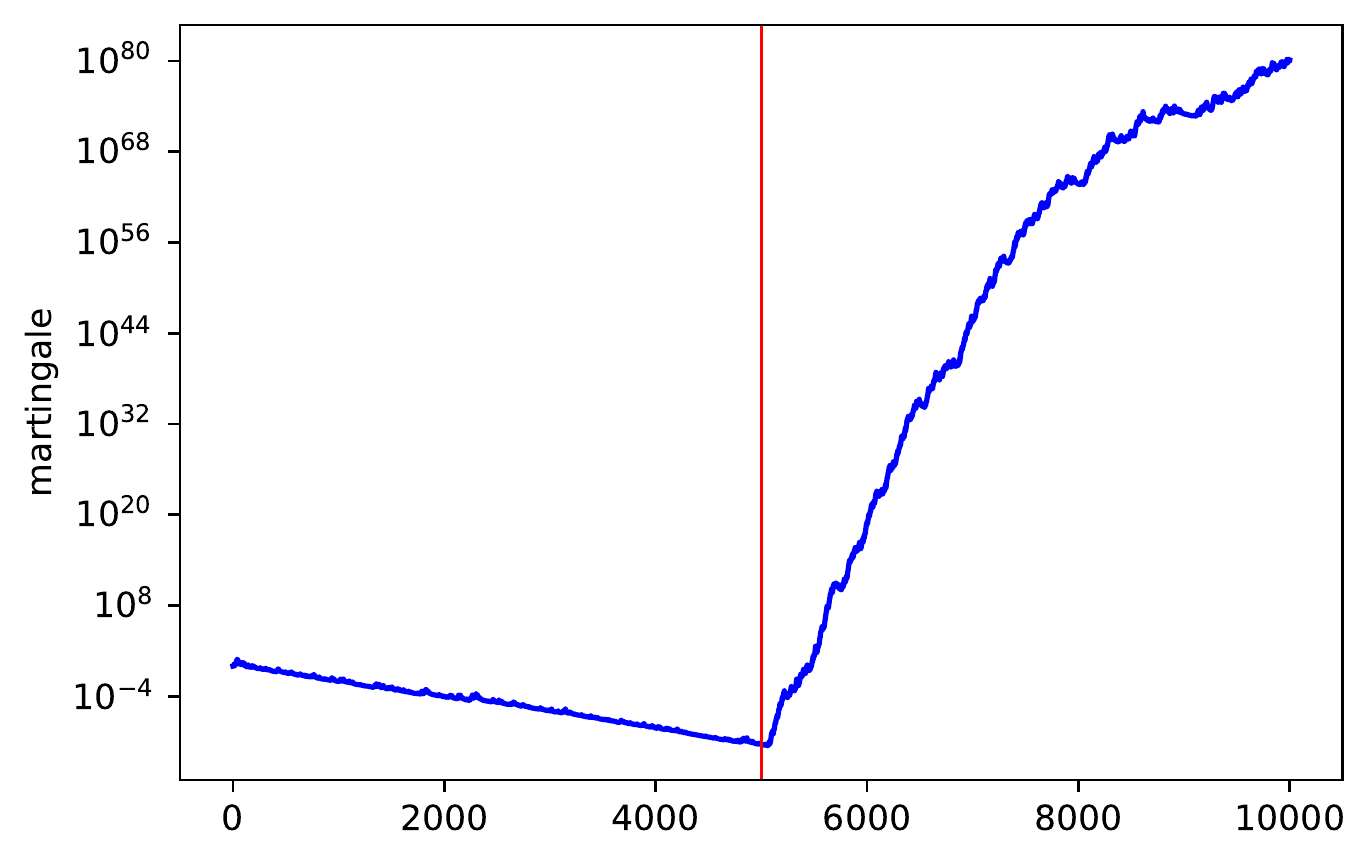}
  \end{center}
  \caption{The safe e-value of \cite{Ramdas/etal:arXiv2102} and the Simple Jumper martingale of \cite{\OCMXXXII},
    as described in text.}
  \label{fig:RRLK}
\end{figure}

The model situation considered in this note is the one chosen by Ramdas et al.\ \cite{Ramdas/etal:arXiv2102}
for studying their process $R_t$, which they call a safe e-value.
Our data consist of binary observations generated independently from Bernoulli distributions.
Let $B(\pi)$ be the Bernoulli distribution on $\{0,1\}$ with parameter $\pi\in[0,1]$:
$B(\pi)(\{1\})=\pi$.
We assume that the observations are IID except that at some point the value of the parameter changes.
Let $\pi_0$ be the pre-change parameter and $\pi_1$ be the post-change parameter.
The total number of observations is $N$,
of which the first $N_0$ come from the pre-change distribution $B(\pi_0)$
and the remaining $N_1:=N-N_0$ from the post-change distribution $B(\pi_1)$.

Ramdas et al.'s \cite{Ramdas/etal:arXiv2102} setting is where $\pi_0=0.1$, $\pi_1=0.4$, $N=10^4$, and $N_0=N_1=5000$.
The plot in the left panel of Figure~\ref{fig:RRLK} is reproduced from their paper
and shows the trajectory of the safe e-value $R_t$.
The right panel of Figure~\ref{fig:RRLK} gives the trajectory
of the Simple Jumper conformal test martingale,
as defined in \cite{\OCMXXXII},
based on the identity conformity measure;
the martingale (including the parameter $J=0.01$)
is exactly as described in \cite[Algorithm 1]{\OCMXXXII}.
Every test martingale is a safe e-value,
and so the conformal test martingale performs surprisingly well.

The goal of this note is to explore attainable final values of test martingales.
Our null hypothesis is the \emph{IID model},
under which the observations are IID but the value of the parameter $\pi$ is unrestricted.

\section{Optimal martingales}
\label{sec:optimal}

In this section we will discuss martingales satisfying various properties of optimality,
not always stated precisely.
Throughout the note,
we use $B(0.1)$ as the pre-change distribution and $B(0.4)$ as the post-change distribution.
For each $n\in\{1,2,\dots\}$, let $k(n)$ be the number of 1s among the first $n$ observations
in the data sequence.

\begin{figure}
  \begin{center}
    \includegraphics[width=0.48\textwidth]{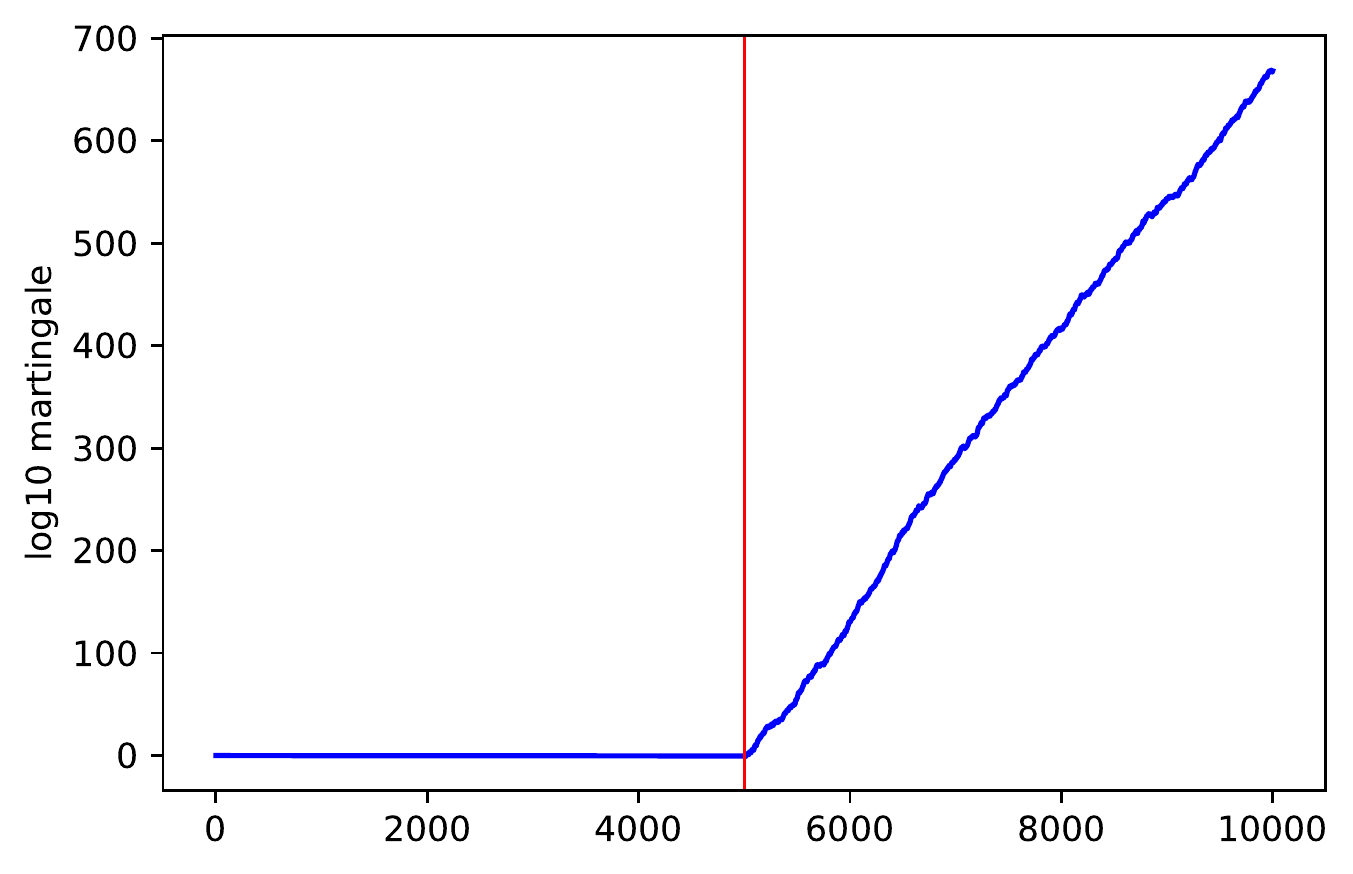}
    \includegraphics[width=0.48\textwidth]{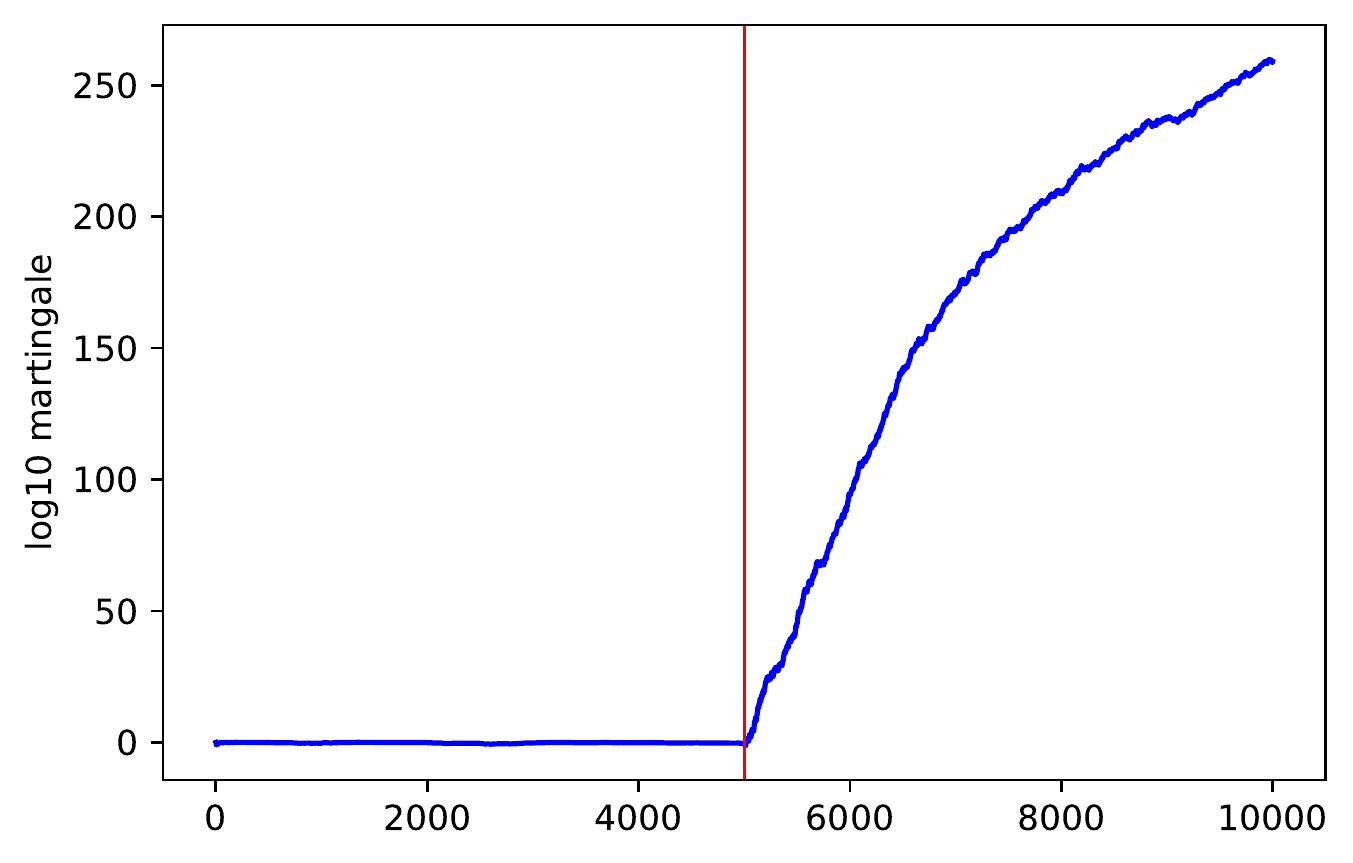}
  \end{center}
  \caption{The likelihood ratio and inf likelihood ratio processes, as described in text;
    the final value in the right panel is approximately $10^{258.93}$.
    In this and the next two figures the vertical axis shows the base 10 logarithm of the martingale value.}
  \label{fig:oracle}
\end{figure}

\begin{figure}
  \begin{center}
    \includegraphics[width=0.48\textwidth]{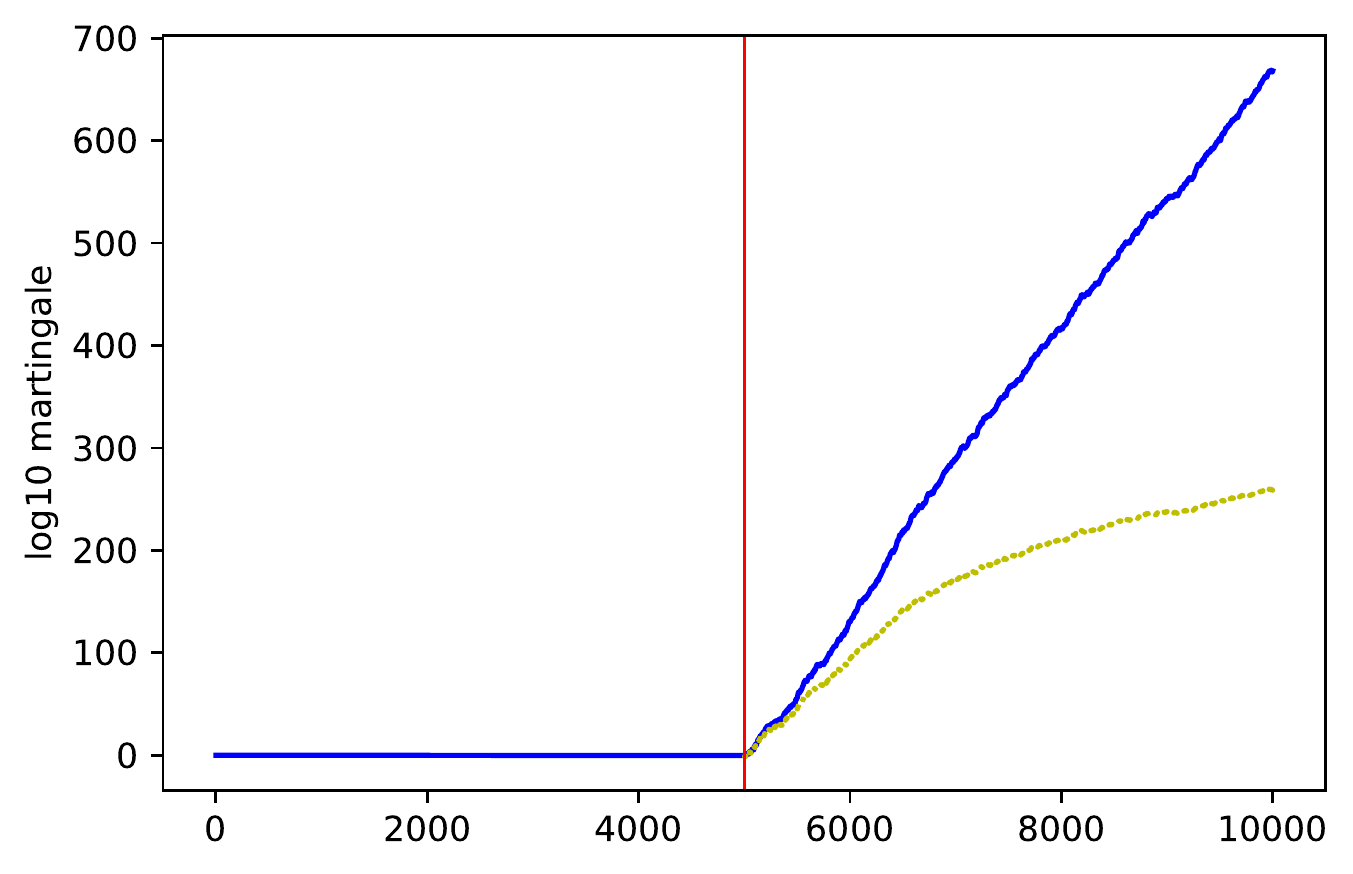}
    \includegraphics[width=0.48\textwidth]{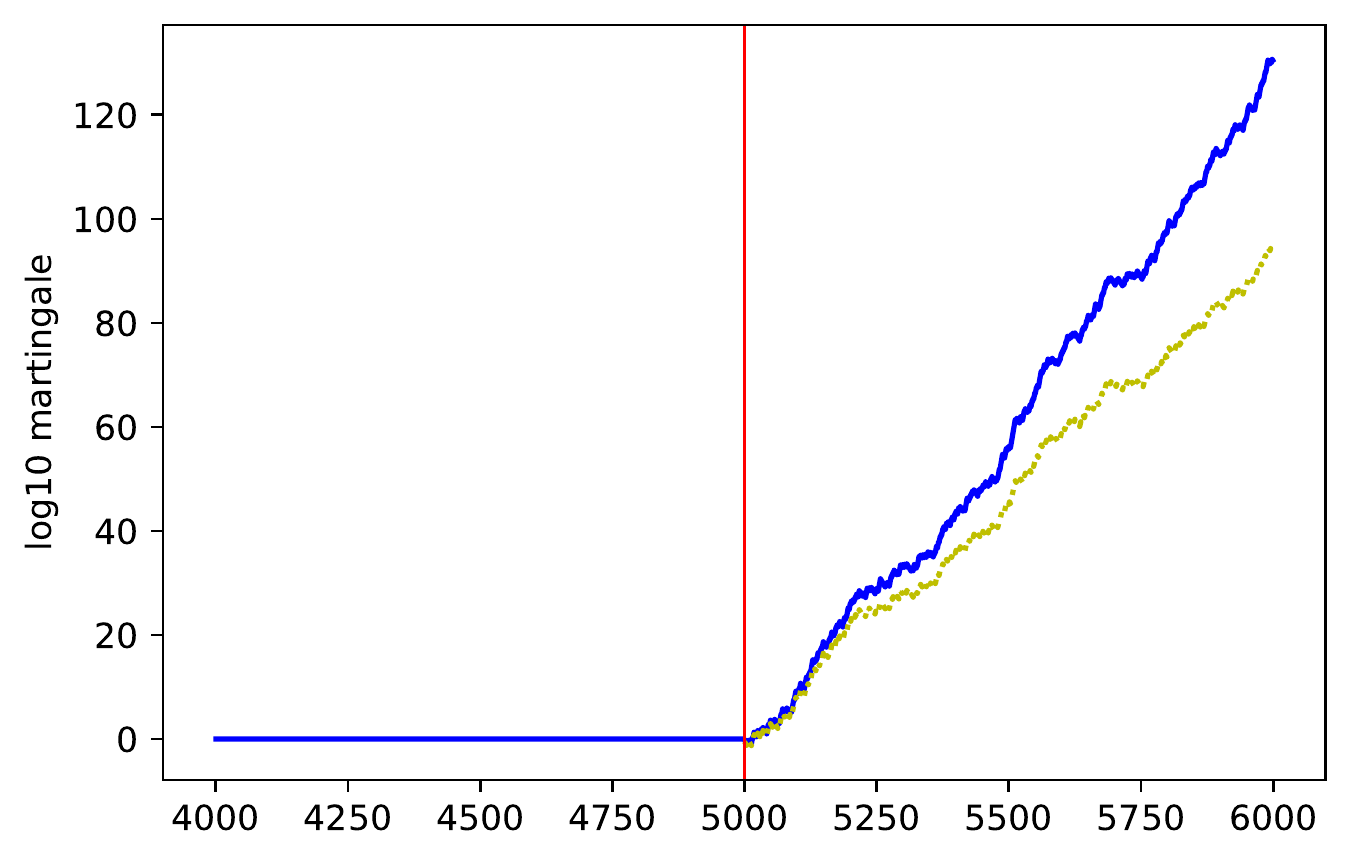}
  \end{center}
  \caption{Left panel: likelihood ratio (solid blue line) and inf likelihood ratio (dotted yellow line)
    over the whole dataset.
    Right panel: over the middle 2000 observations.}
  \label{fig:comparison}
\end{figure}

\begin{figure}
  \begin{center}
    \includegraphics[width=0.48\textwidth]{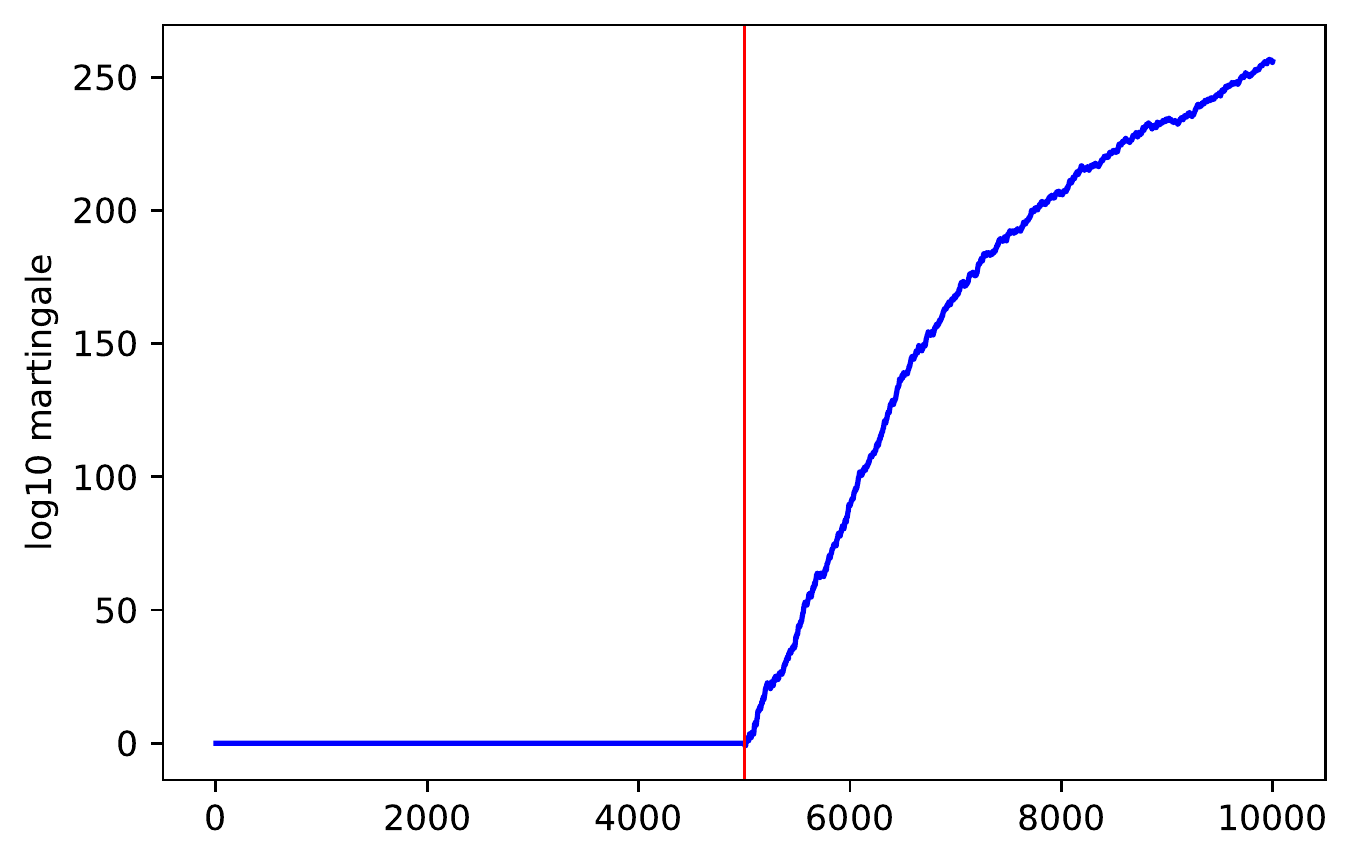}
    \includegraphics[width=0.48\textwidth]{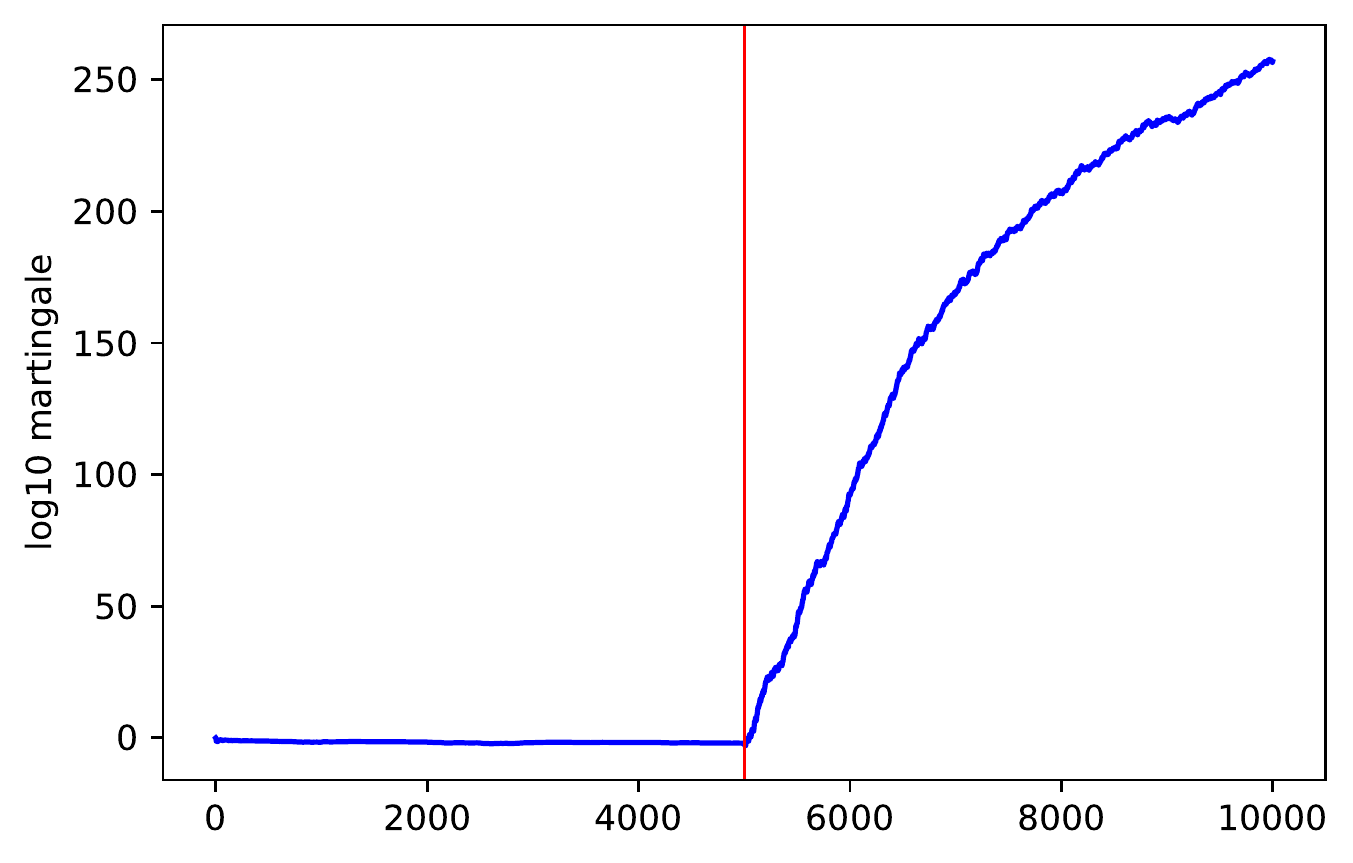}
  \end{center}
  \caption{The optimal conformal test martingale and the corresponding conformal e-pseudomartingale, as described in text;
    their final values are approximately $10^{255.84}$ and $10^{256.89}$, respectively.}
  \label{fig:conformal}
\end{figure}

The left panel of Figure~\ref{fig:oracle} shows the trajectory
of the likelihood ratio of the true distribution
to the pre-change distribution extended to the full dataset:
\[
  S^{(0)}_n
  :=
  \begin{cases}
    1 & \text{if $n\le N_0$}\\
    \left(\frac{\pi_1}{\pi_0}\right)^{k(n)-k(N_0)}
    \left(\frac{1-\pi_1}{1-\pi_0}\right)^{(n-N_0)-(k(n)-k(N_0))}
    & \text{otherwise}.
  \end{cases}
\]
This is an optimal test martingale in Wald's \cite{Wald:1947,Wald/Wolfowitz:1948} sense.
This process, however, is a test martingale only with respect to the null hypothesis $B(0.1)$,
whereas our null hypothesis is the IID model.
The right panel of Figure~\ref{fig:oracle} shows the infimum of the likelihood ratios
\begin{equation}\label{eq:S1}
  S^{(1)}_n
  :=
  \begin{cases}
    \frac{\pi_0^{k(n)} (1-\pi_0)^{n-k(n)}}{\left(\frac{k(n)}{n}\right)^{k(n)}\left(1-\frac{k(n)}{n}\right)^{n-k(n)}}
    & \text{if $n\le N_0$}\\[3mm]
    \frac
      {\pi_0^{k(N_0)} (1-\pi_0)^{N_0-k(N_0)} \pi_1^{k(n)-k(N_0)} (1-\pi_1)^{(n-N_0)-(k(n)-k(N_0))}}
      {\left(\frac{k(n)}{n}\right)^{k(n)}\left(1-\frac{k(n)}{n}\right)^{n-k(n)}}
    & \text{otherwise},
  \end{cases}
\end{equation}
where $0^0:=1$.
We will refer to this process as the \emph{inf likelihood ratio};
its final value is indicative of the best result that can be attained in our testing problem.
Notice that in Figures~\ref{fig:oracle}--\ref{fig:conformal}
the vertical axis shows the base 10 logarithm of the martingale value.

\begin{remark}\label{rem:difficulty}
  The expression \eqref{eq:S1} is the infimum over the IID measures
  of the likelihood ratios that are individually optimal (for each IID measure)
  in Wald's sense.
  However, this does not necessarily mean that the infimum \eqref{eq:S1} itself is optimal.
  The extreme case is where the null hypothesis consists of all probability measures on $\{0,1\}^{\infty}$.
  The inf likelihood ratio will quickly tend to 0,
  and so its performance will be much worse than that of the identical 1
  (which is a test martingale).
  The case of \eqref{eq:S1}, however, is very far from this extreme case,
  and even to the left of $N_0$ the trajectory of $\log_{10} S^{(1)}$ is visually indistinguishable from zero.
\end{remark}

Figure~\ref{fig:comparison} compares the two processes shown in Figure~\ref{fig:oracle}.
The likelihood ratio process grows exponentially fast,
which shows as a linear growth on the log scale.
Its trajectory looks like a tangent to the inf likelihood ratio trajectory.
It is clear that the inf likelihood ratio cannot grow exponentially fast:
the post-change distribution $B(0.4)$ is gradually becoming ``the new normal''.

Next let us find the optimal conformal test martingale,
optimized under the true data distribution.
For the definition of conformal test martingales,
see, e.g., \cite{\OCMXXIV,\OCMXXXII}.
During the first $N_0$ trials we do not gamble,
so let us consider a trial $n>N_0$.
Taking the identity function as the nonconformity measure
(the difference between conformity and nonconformity is essential in this context),
we obtain a p-value $p_n \in [0,k(n)/n]$ with probability $\pi_1$,
and we obtain $p_n \in [k(n)/n,1]$ with probability $1-\pi_1$.
Since the expected value of $k(n)/n$ is $(N_0 \pi_0 + (n-N_0)\pi_1)/n$,
the likelihood ratio betting function
\begin{equation}\label{eq:optimal}
  f_n(p)
  :=
  \begin{cases}
    \frac{n \pi_1}{N_0 \pi_0 + (n-N_0)\pi_1} & \text{if $p\le\frac{N_0 \pi_0 + (n-N_0)\pi_1}{n}$}\\[1mm]
    \frac{n (1-\pi_1)}{N_0 (1-\pi_0) + (n-N_0)(1-\pi_1)} & \text{otherwise}
  \end{cases}
\end{equation}
is optimal, in some sense,
as shown in \cite[Theorem 2]{\OCMIV}.
The left panel of Figure~\ref{fig:conformal} shows the trajectory
of the corresponding conformal test martingale.

The betting functions \eqref{eq:optimal} involve the expected value of $k(n)/n$.
We can slightly improve the performance of the conformal test martingale
shown in the left panel of Figure~\ref{fig:conformal} if we replace \eqref{eq:optimal} by
\begin{equation*}
  f_n(p)
  :=
  \begin{cases}
    \frac{n \pi_1}{k(n)} & \text{if $p\le\frac{k(n)}{n}$}\\[1mm]
    \frac{n (1-\pi_1)}{n-k(n)} & \text{otherwise}.
  \end{cases}
\end{equation*}
However, the resulting process is not a genuine martingale but a conformal e-pseudomartingale,
in the terminology of \cite{\OCMXXIX}.
It is shown in the right panel of Figure~\ref{fig:conformal}.

It is interesting that the right panel of Figure~\ref{fig:oracle}
and the left and right panels of Figure~\ref{fig:conformal}
are visually almost indistinguishable,
but in fact the final value for the conformal e-pseudomartingale is about 10 times larger
than the final value of the conformal test martingale,
and the final value of the inf likelihood ratio is in turn more than 100 times larger
than the final value of the conformal e-pseudomartingale.

\section{A more natural conformal martingale}

The martingales whose trajectories are shown in Figures~\ref{fig:oracle}--\ref{fig:conformal}
depend very much on the knowledge of the true data-generating mechanism.
Can we obtain comparable results without blatant optimization (requiring such knowledge)?
This is the topic of this section.

Let us generalize the betting function \eqref{eq:optimal} to
\begin{equation}\label{eq:LR}
  f_{a,b}(p)
  :=
  \begin{cases}
    \frac{b}{a} & \text{if $p\le a$}\\[1mm]
    \frac{1-b}{1-a} & \text{otherwise},
  \end{cases}
\end{equation}
where $a,b\in(0,1)$.
It is easy to see that $\int f_{a,b} = 1$.
Apart from the betting functions \eqref{eq:LR} we will use the identity function $f_{\square}$,
$f_{\square}(p):=p$.
Let $S_n$ be the conformal test martingale
\begin{equation}\label{eq:S}
  S_n
  :=
  \int f_{x_1}(p_1) \dots f_{x_n}(p_n) \mu(\ddd(x_1,x_2,\dots)),
\end{equation}
where $p_1,p_2,\dots$ is the underlying sequence of conformal p-values
and $\mu$ is the distribution of the following Markov chain.

The Markov chain is defined in the spirit of tracking the best expert in prediction with expert advice.
The state space is $\{\square\}\cup(0,1)^2$,
and $R\in(0,1)$ is the parameter (typically a small number).
The initial state is $\square$ (the \emph{sleeping} state).
The transition function is:
\begin{itemize}
\item
  if the current state is $\square$,
  with probability $1-R$ the state remains $\square$,
  and with probability $R$ a new state $(a,b)$ is chosen from the uniform distribution in $(0,1)^2$;
\item
  the states $(a,b)\in(0,1)^2$ are absorbing:
  if the current state is $(a,b)\in(0,1)^2$, it will stay $(a,b)$.
\end{itemize}

\begin{algorithm}[bt]
  \caption{Sleeper/Chooser ($(p_1,p_2,\dots)\mapsto(S_1,S_2,\dots)$)}
  \label{alg:SC}
  \begin{algorithmic}[1]
    \State $S_{\square}:=1$
    \For{$(a,b)\in\mathbf{G}^2$:}
      $S_{a,b}:=0$
    \EndFor
    \For{$n=1,2,\dots$:}
      \For{$(a,b)\in\mathbf{G}^2$:}
        $S_{a,b} := S_{a,b} f_{a,b}(p_n)$
      \EndFor
      \State $S_n := S_{\square} + \sum_{(a,b)\in\mathbf{G}^2} S_{a,b}$
      \For{$(a,b)\in\mathbf{G}^2$:}
        $S_{a,b} := S_{a,b} + R S_{\square} / (G-1)^2$ \label{ln:share-1}
      \EndFor
      \State $S_{\square} := (1-R)S_{\square}$ \label{ln:share-2}
    \EndFor
  \end{algorithmic}
\end{algorithm}

In our implementation of the procedure \eqref{eq:S},
we replace the square $(0,1)^2$ by the grid $\mathbf{G}^2$,
where
\[
  \mathbf{G}
  :=
  \left\{
    \frac1G, \frac2G,\dots,\frac{G-1}{G}
  \right\}
\]
and $G$ (positive integer) is another parameter.
The resulting procedure is shown as Algorithm~\ref{alg:SC}.

The intuition behind Algorithm~\ref{alg:SC} is that,
in order to gamble against the uniformity of $(p_1,p_2,\dots)$,
we distribute our initial capital of 1 among accounts $S_{a,b}$ indexed by $(a,b)\in\mathbf{G}^2$,
and there is also a sleeping account $S_{\square}$.
We start from all money invested in the sleeping account,
but at the end of each step a fraction $R$ of that money is moved to the active accounts $S_{a,b}$
and divided between them equally (see lines \ref{ln:share-1} and \ref{ln:share-2}).
On account $S_{a,b}$ we gamble against the uniformity of the input p-values
using the calibrator $f_{a,b}$.

\begin{figure}
  \begin{center}
    \includegraphics[width=0.6\textwidth]{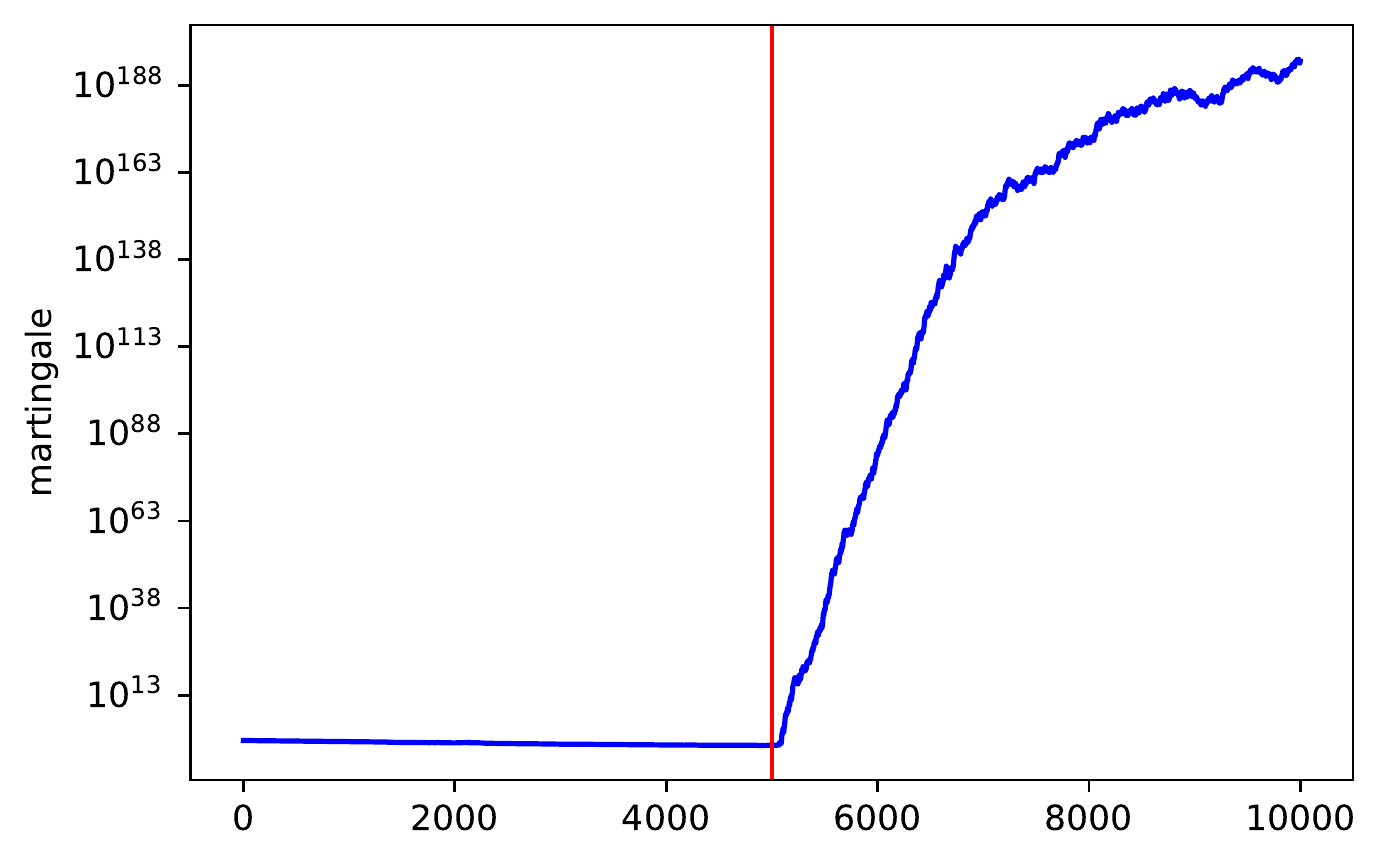}
  \end{center}
  \caption{The basic conformal test martingale inspired by the optimal one, as described in text;
    the final value is approximately $7.85 \times 10^{194}$.}
  \label{fig:base}
\end{figure}

Figure~\ref{fig:base} suggests that we can improve on the result of Figure~\ref{fig:RRLK}
using a fairly natural, and in fact very basic, conformal test martingale.
We use the identity conformity measure and the Sleeper/Chooser betting martingale of Algorithm~\ref{alg:SC},
and the parameters are $R:=0.001$ and $G:=100$;
therefore, $a$ and $b$ are chosen from the grid $\{0.01,0.02,\dots,0.99\}$.
The final value of the resulting conformal test martingale is closer (on the log scale)
to those in Figure~\ref{fig:conformal} than in Figure~\ref{fig:RRLK}.

\section{Conclusion}

In this note we have discussed only the case of binary observations,
in which the simple calibrators \eqref{eq:LR} are appropriate.
This can be regarded as first step of an interesting research programme.
We can simulate different model situations that can be analyzed theoretically
and develop suitable conformal test martingales,
as we did in this note for a binary model situation.
Perhaps the next in line are the Gaussian model with a constant variance and a change in the mean,
the Gaussian model with a constant mean and a change in the variance,
and the exponential model
(as in, e.g., \cite[Part II]{Wald:1947} and \cite{Tartakovsky/etal:2015}).
Optimal conformal test martingales (such as those in Section~\ref{sec:optimal})
provide a clear goal for more natural conformal test martingales,
and even give ideas of how this goal can be attained.
These ideas, in turn, add to the toolbox that we can use for dealing with practical problems,
where we often have only a vague notion of the true data-generating distribution.
One difficulty, however, is that the IID model in these cases becomes very large,
and the complication alluded to in Remark~\ref{rem:difficulty} becomes more worrisome.

\subsection*{Acknowledgments}

This work has been supported by Amazon and Stena Line.

\end{document}